\documentclass{article}

\usepackage{PRIMEarxiv}

\usepackage[utf8]{inputenc} 
\usepackage[T1]{fontenc}    
\usepackage{hyperref}       
\usepackage{url}            
\usepackage{booktabs}       
\usepackage{amsfonts}       
\usepackage{nicefrac}       
\usepackage{microtype}      
\usepackage{lipsum}
\usepackage{fancyhdr}       
\usepackage{graphicx}       
\graphicspath{{img/}}     

\usepackage{graphicx}
\usepackage{cuted}
\usepackage{multirow}
\usepackage{mathtools}
\usepackage{caption}
\usepackage{amsmath}
\usepackage{relsize}
\usepackage{algorithm}
\usepackage{algpseudocode}
\usepackage{siunitx}
\usepackage{orcidlink}
\usepackage{authblk}
\usepackage{amssymb}

\usepackage{pifont}
\newcommand{\cmark}{\ding{51}}%
\newcommand{\xmark}{\ding{55}}%

\usepackage[symbol]{footmisc}
  
\title{EPOCH: Jointly Estimating the 3D Pose of Cameras and Humans 
}

\author[1,2]{Nicola Garau\textsuperscript{$\dag$}}
\author[1]{Giulia Martinelli}
\author[1]{Niccolò Bisagno}
\author[2]{Denis Tomè}
\author[2]{Carsten Stoll}

\affil[1]{University of Trento}
\affil[2]{Epic Games}
\affil[ ]{\texttt{nicola.garau@unitn.it}}

\begin{document}
\maketitle

\footnotetext[2]{Work primarily done during an internship at Epic Games.}


\begin{abstract}
    Monocular Human Pose Estimation (HPE) aims at determining the 3D positions of human joints from a single 2D image captured by a camera. However, a single 2D point in the image may correspond to multiple points in 3D space. Typically, the uniqueness of the 2D-3D relationship is approximated using an orthographic or weak-perspective camera model.
    In this study, instead of relying on approximations, we advocate for utilizing the full perspective camera model. This involves estimating camera parameters and establishing a precise, unambiguous 2D-3D relationship.
    
    To do so, we introduce the EPOCH framework, comprising two main components: the pose lifter network (LiftNet) and the pose regressor network (RegNet). LiftNet utilizes the full perspective camera model to precisely estimate the 3D pose in an unsupervised manner. It takes a 2D pose and camera parameters as inputs and produces the corresponding 3D pose estimation. These inputs are obtained from RegNet, which starts from a single image and provides estimates for the 2D pose and camera parameters. RegNet utilizes only 2D pose data as weak supervision. Internally, RegNet predicts a 3D pose, which is then projected to 2D using the estimated camera parameters. This process enables RegNet to establish the unambiguous 2D-3D relationship.

    Our experiments show that modeling the lifting as an unsupervised task with a camera in-the-loop results in better generalization to unseen data. We obtain state-of-the-art results for the 3D HPE on the Human3.6M and MPI-INF-3DHP datasets. Our code is available at: [Github link upon acceptance, see supplementary materials].
\end{abstract}

\begin{figure}
    \centering
    \includegraphics[width=0.6\linewidth]{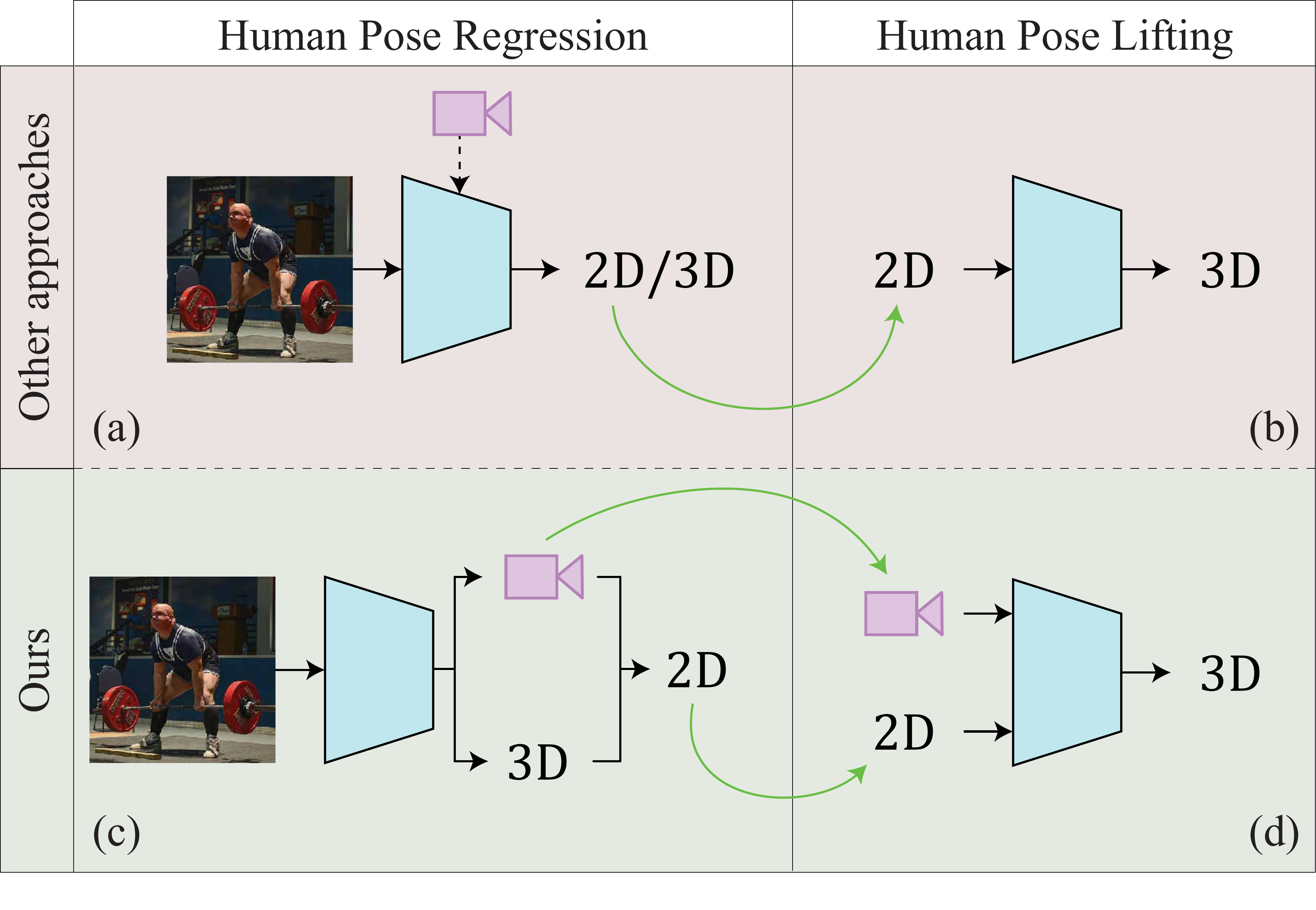}
    \caption{ (a) In human pose estimation, classical approaches perform a direct regression of the 2D/3D joint location directly from an image. If the ground truth is available, the camera parameters can be used/learned to refine the accuracy. (b) Lifting approaches aim at retrieving the depth of each 2D joint to obtain the 3D pose. (c) We propose a novel paradigm, \textit{directly estimating the 3D pose and the camera from images.} The 2D pose can be calculated by applying the projection of the 3D coordinates to the image space using the camera parameters. (d) \textit{Starting from the estimated 2D poses and camera parameters, we perform the lifting to 3D}, improving the performances with respect to current approaches.}
    \label{fig:teaser}
\end{figure}

\section{Introduction}
\label{sec:introduction}

There are two main approaches to monocular 3D human pose estimation (HPE) from a single RGB images~\cite{ji2020survey}. One class of algorithms uses a single-stage approach, where the aim is to regress the 3D position of human joints directly from an image \cite{pavlakos2018ordinal,pavlakos2017coarse,sun2017compositional}. The other class of approaches use two distinct stages, where the first step is to infer 2D poses from monocular RGB images, which is followed by a lifter network that predicts the 3D displacement for each of the 2D joints. Two stage approaches typically outperform single stage approaches \cite{zheng2023deep}.

Estimating 3D human poses from a single RGB image is difficult as the problem is inherently \textit{ill-posed}. For any 2D observation there exist multiple plausible 3D poses that will lead to the same 2D projection~\cite{zhang2022survey,SMPL:2015}. Additionally, collecting reliable ground-truth 3D data is difficult. Annotating 3D ground truth on 2D images inevitably introduces inaccuracies, and collecting actual ground truth requires a complex and expensive controlled environment using multi-view camera systems or additional capture modalities. Even using multiple views, triangulation can lead to errors, as there are inherent ambiguities in the position of the joint under the body surface. While limited datasets providing 3D data are available~\cite{ionescu2013human3, joo2015panoptic}, 2D datasets still provide more data in more general scenarios and environments.

To address the \textit{ill-posed} nature of the problem, past approaches have relied on different strategies, like fully supervised training using either real \cite{dabral2018learning,shan2023diffusion} or synthetic 3D ground truth \cite{kundu2022uncertainty,kundu2021non}, weakly supervised training relying on multiple views:~either paired \cite{usman2022metapose,wandt2021canonpose} or unpaired \cite{wandt2019repnet}, 2D supervision \cite{tripathi2020posenet3d}, or video motion consistency \cite{huang2022occluded,gholami2022adaptpose,hu2021unsupervised}. Unsupervised approaches have relied on cycle consistency coupled with a weak perspective camera projection to lift 2D poses to 3D \cite{wandt2022elepose,chen2019unsupervised}. Relying on a weak perspective camera projection is not ideal because it does not accurately capture the perspective transformation \cite{hartley2003multiple}, leading to depth inaccuracy and scale ambiguity when projecting a 2D skeleton in the 3D space. Recent works \cite{kocabas2021spec} have shown that using fully perspective cameras reduces this ambiguity.


In this paper, we introduce EPOCH, a novel unsupervised framework that effectively addresses the challenges of \textit{data scarcity} by harnessing unsupervised techniques and mitigates the inherent \textit{ill-posed nature} of the problem through explicit camera modeling, as shown in Fig.~\ref{fig:teaser}. Our approach stands out due to its capability to estimate the full perspective camera parameters leveraging only 2D human poses, without relying on any camera ground truth. We claim that by incorporating the estimated camera into the 3D lifting operation, it is possible to enhance the accuracy and the consistency of 3D unsupervised human pose estimation while generalizing to unseen data. 
Our method consists of an unsupervised 3D human pose lifter network (\textbf{LiftNet}) and a lightweight capsule-based regressor network (\textbf{RegNet}) that estimates the camera pose and 2D joint positions.

\textbf{LiftNet} performs the 3D lifting from estimated 2D poses and camera parameters. Inspired by \cite{chen2019unsupervised}, we employ a self-supervising cycle-consistent framework. Unlike \cite{wandt2022elepose,chen2019unsupervised}, our approach uses a full perspective camera, allowing us to use a wider range of camera transformations for supervision in our cycle consistency, and improving the accuracy of the model.

We estimate the camera pose and 2D poses used as input for the lifting stage using \textbf{RegNet}, a lightweight capsule-based regressor network that is trained on weakly supervised 2D pose data instead of fully supervised data as \cite{kocabas2021spec}.  It use contrastive pre-training and heatmap-free joint position regression to estimate the 2D poses as well as the intrinsic (the camera matrix $[K]$) and extrinsic parameters (rotation matrix $[R]$ and translation vector $T$) of a camera based on the standard pinhole camera model. To supervise the camera estimation with the 2D pose data we internally predict a 3D pose that is then projected to 2D with the estimated camera. The internally predicted 3D pose is not quite as accurate as the refined output of \textbf{LiftNet} but helps regularizing the camera estimation.

With no prior about 3D human appearance, both \textbf{LiftNet} and \textbf{RegNet} estimate a single 2D projection which is not enough to guarantee a plausible 3D pose. Thus, we employ Normalizing Flows (NF) to ensure the plausibility of multiple 2D projections of a single 3D estimate. Different from previous approaches \cite{wandt2022elepose}, our NF is based on simple 1x1 convolutions \cite{kingma2018glow}, which can be applied to the full feature representation of the poses, without the need to reduce their dimensionality using the Principal Component Analysis (PCA).

The EPOCH framework is the sequential combination of \textbf{RegNet} and \textbf{LiftNet}, which allows for the direct inference of accurate 3D poses from images. \textbf{RegNet} estimates 3D poses with weak 2D pose supervision, while the camera parameters are estimated without any ground truth camera data. \textbf{LiftNet} predicts 3D poses based on the estimates of 2D poses and camera poses. We argue that \textbf{RegNet} is weakly supervised, whereas \textbf{LiftNet} is fully unsupervised as it relies solely on estimates, without any ground truth data. This reasoning applies to both poses and cameras.

The novelty of our work can be summarized as:

\begin{itemize}
    \item We define an innovative EPOCH framework to address the challenges of \textit{data scarcity} and the \textit{ill-posed nature} of 3D HPE problem by harnessing the full camera perspective projection, enabling the direct lifting of accurate 3D poses from input images. 
    \item We present \textbf{LiftNet}, a novel 3D unsupervised HPE framework that leverages perspective camera projection to improve the accuracy of 2D pose lifting.
    \item We introduce an original capsule-based regressor, called \textbf{RegNet}, which jointly estimates 3D joint positions and camera parameters. Final 2D joints are computed using the estimated camera by perspective projection.
    \item We adopt a lightweight Normalizing Flows \cite{kingma2018glow} model to enforce anthropomorphic constraints. Our NF accepts the entire 2D skeleton as input without the need for dimensionality reduction using PCA.
    \item We obtain state-of-the-art results on both 3D HPE direct regression and 3D unsupervised HPE on the common benchmark datasets Human3.6M and MPI-INF-3DHP.
\end{itemize}

\section{Related work}
\label{sec:related_work}

3D human pose estimation (HPE) from monocular 2D images has been extensively researched through supervised, weakly-supervised, and unsupervised approaches \cite{zheng2023deep,liu2022recent}. This section gives an overview of the different methods, also focusing on the challenging in-the-wild approaches.

\textbf{Fully supervised approaches.} In this paradigm the 3D ground truth is readily available. Gathering such data requires collecting vast datasets such as Human3.6M \cite{ionescu2013human3}, 3DPW \cite{von2018recovering}, MPI-INF-3DHP \cite{mono-3dhp2017} and CMU Panoptic \cite{joo2015panoptic}. 
In supervised methods, two primary strategies exist: direct regression of 3D coordinates from the image \cite{pavlakos2018ordinal,pavlakos2017coarse,sun2017compositional} (Fig.~\ref{fig:teaser}(a)), or 2D pose estimation followed by lifting to 3D \cite{xu2022vitpose,yang2021transpose,zhu2023motionbert,chun2023learnable} (Fig.~\ref{fig:teaser}(b)).
Direct regression proves more challenging because it involves the simultaneous estimation of 3D coordinates for each joint, often leading to inferior results compared to lifting-based techniques \cite{zheng2023deep}. 
Supervised networks achieve the best results on multiple datasets \cite{zhu2023motionbert,chun2023learnable}, but often struggle to generalize to different scenarios like out-of-distribution poses, challenging camera angles and in-the-wild pose estimation \cite{zheng2023deep}.

\textbf{Weakly-supervised approaches} rely on the lifting framework \cite{usman2022metapose,wandt2021canonpose,wandt2019repnet, tripathi2020posenet3d,hu2021unsupervised,zhang2022mixste,li2022mhformer,kundu2020self,drover2018can} using various supervision signals without directly accessing the 3D ground truth paired with the corresponding 2D image. For instance, multiple paired or unpaired views of the same subject provide a supervision signal, through the consistency of the estimated 3D pose seen from different viewpoints \cite{usman2022metapose,wandt2021canonpose,wandt2019repnet,rhodin2018learning,kundu2020self,kocabas2019self}.
In monocular approaches, temporally correlated 2D poses can be estimated from an input video and used as a supervision signal for a frame-specific 3D pose estimation \cite{hu2021unsupervised,zhang2022mixste,li2022mhformer}. In-the-wild approaches have mostly relied on 2D pose as ground truths to supervise intermediate 3D estimates \cite{yu2021towards,habibie2019wild}. Other approaches perform monocular 3D pose estimation using only 2D pose supervision \cite{kundu2020self,fish2017adversarial,chen2019unsupervised,wandt2019repnet}.

Following this line of work, our regressor network (\textbf{RegNet}) is a novel weakly-supervised approach that uses 2D poses for supervision, computing the 3D to 2D projection via estimated camera parameters. Moreover, our approach is the first that jointly estimates the full perspective camera parameters without relying on any ground truth camera.

\textbf{Unsupervised approaches.} Unsupervised approaches usually rely on the lifting paradigm, employing multiple different signals to regularise their 3D predictions. In~\cite{chen2019unsupervised}, the authors proposed an unsupervised lifting network grounded in closure and invariance properties, incorporating a geometric self-consistency loss. The closure property for a lifted 3D skeleton means that, after random rotation and re-projection, the resulting 2D skeleton will lie within the distribution of valid 2D poses. The invariance propriety means that, when changing the viewpoint of 2D projection from a 3D skeleton, the re-lifted 3D skeleton should be the same. Following a similar concept \cite{wandt2022elepose} introduces a weak camera projection to model the lift-reproject-lift process. The weak camera projection is coupled with the elevation estimation of the camera, providing an approximation of the full camera perspective model. Moreover, they introduce the use of Normalizing Flows (NFs) \cite{dinh2016density}, which are used to ensure the closure property more accurately than GAN-based methods \cite{chen2019unsupervised}. 
In \cite{hardy2023unsupervised} a similar framework is extended to a multi-person scenario, where relative positions are also used as supervision signals.

Our lifter network (\textbf{LiftNet}) follows this line of work while introducing the following novelties: (i) we employ a full perspective camera model for the projection, making it much more accurate and robust to varying focal lengths, (ii) we drop the need for an additional elevation prediction branch in the lifting network \cite{wandt2022elepose}, (iii) we avoid applying the PCA to reduce data dimensionality by training a normalizing flow based on Glow \cite{kingma2018glow} instead of RealNVP, (iv) we add a geometric constraint on unnatural joints folds.

\section{Method}
\label{sec:method}


\subsection{Preliminaries}
\label{sec:preliminaries}

\textbf{Camera model.}
While many prior works for 3D human pose estimation rely on simplified weak perspective camera models, we use a full perspective camera model, consisting of \textit{intrinsic} parameters \textbf{K} (focal length and center of projection) and \textit{extrinsic} parameters \textbf{R} and \textbf{T} (rotation and translation of the camera respectively). We transform a 3D point $(X, Y, Z)$ into camera coordinates by multiplying with the \textit{extrinsic} and \textit{intrinsic} matrices, and get the final image space coordinates $I = [u/w, v/w]$. 

\begin{equation}
\label{eq:full_camera_reprojection}
    \begin{bmatrix}
        u \\ v \\ w
    \end{bmatrix}_{2D}
    =
    \mathlarger{[ \textbf{K}]}
    \mathlarger{[ \textbf{R} | \textbf{T}]}     
    \begin{bmatrix}
        X \\ Y \\ Z \\ 1
    \end{bmatrix}_{3D}
\end{equation}

To invert the projection given image coordinates $\Tilde{I}$, we need to estimate the unknown depth $w$ of the point.
Similar to \cite{li2022cliff} we estimate the intrinsic parameters directly from the full image size and the bounding box crop using a model $\Psi$. While this is only an approximation of the real field-of-view of the camera, prior work \cite{Kissos2020BeyondWP} has shown this to be a good approximation for most real life cameras used. The estimated intrinsic camera matrix $[K]$ includes the focal length $f=(f_w, f_h)$ and the principal point $c=(c_w,c_h)$. Moreover, we estimate a scaling factor $s=(s_w,s_h)$ for the regularisation of the image size and skeleton height. See Supplementary Materials for further details on model $\Psi$.

\textbf{Normalizing flows loss.}
Normalizing flows (NFs) are a class of generative neural networks capable of mapping a complex distribution into a simpler one using invertible functions \cite{kobyzev2020normalizing}. They are trained to learn the probability density function (PDF) of a given dataset relying on an invertible function $f$. See Supplementary material for further details about the modeling of this function. When presented with a novel sample, NFs can estimate the likelihood (plausibility) that the given sample belongs to the learned dataset distribution.




In \cite{wandt2022elepose}, the learnable function $f$ is based on RealNVP \cite{dinh2016density} which is not suitable for high dimensional data like 2D poses, necessitating a PCA reduction of the input vector $x$ for an optimal convergence during training. In contrast, our NFs are based on the Glow framework \cite{kingma2018glow} relying on 1x1 convolutions. Small and fast convolutions allow the use of the full 2D joints' coordinates without the need for a feature reduction as well as reducing the computational costs.

During the training of our architecture, the NFs are used to verify whether multiple projections of 3D poses are all plausible 2D poses without relying on multi-view data. To achieve this, we define the normalizing flow loss $\mathcal{L}_{NF}$, as the negative log-likelihood of the PDF:
\begin{equation}
    \mathcal{L}_{NF}(x) = -log(p_x(\mathbf{x}))
\end{equation}

\begin{figure}
    \centering
    \includegraphics[width=0.65\linewidth]{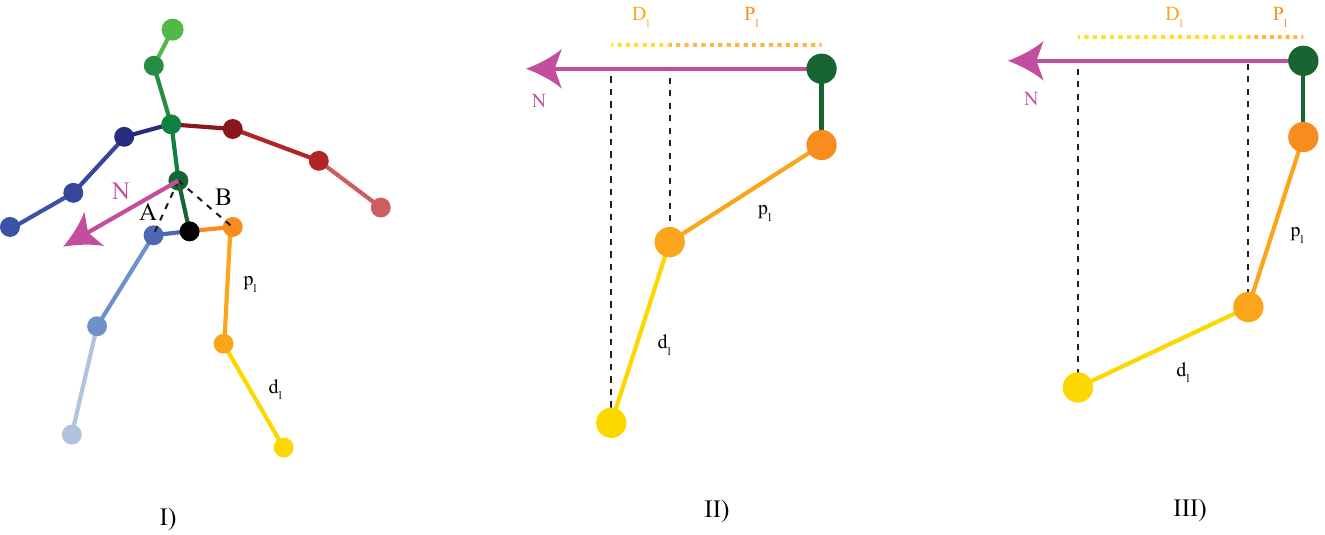}
    \caption{In (I), we define two vectors, denoted as $A$ and $B$, connecting the spine and the hip joints. The cross product of these vectors yields the normal vector $N$, which aligns with the walking direction. In (II) and (III), we show the outcome of the dot product between $N$ and the proximal $p_l$ and distal $d_l$ components, resulting in their projections $D_l$ and $P_l$. In (II), $\mathcal{L}_{limbs}$ gives an output of 0, indicating a anthropomorphically complaint prediction. In (III), $\mathcal{L}_{limbs}$ returns a positive value, signaling the need for further correction.}
    \label{fig:limbs_loss}
\end{figure}

\textbf{Anthropomorphic constraints.}
In supervised 3D HPE, the neural network has explicit access to the 3D ground truth to learn the appearance of a 3D human pose.
In unsupervised or semi-supervised settings, we introduce regularising losses to ensure that the estimated 3D poses respect anthropomorphic constraints, such as proportional bone lengths and articulation angle limits.

As in \cite{wandt2022elepose}, we use a bones ratio loss $\mathcal{L}_{bone}(y)$ to ensure that the ratio between bones lengths of 3D pose $y \in \mathbb{R}^{3J}$ are respected. This loss leverages the observed nearly constant ratio between bones across different individuals \cite{pietak2013fundamental}, without fixing the bone length to a pre-defined value.

Additionally, we define a novel $\mathcal{L}_{limbs}$ loss which ensures that joints like knees and elbows do not bend in unrealistic manners (e.g.\ facing backward with respect to the normal walking direction). It is defined as:
\begin{equation}\
    \mathcal{L}_{limbs}(y) = \frac{1}{L}\sum_l^L{max(0, P_l - D_l)}
\label{eq:limbs_loss}
\end{equation}
where $L$ represents the number of limbs, $P_l = N \cdot p_l$ and $D_l = N \cdot d_l$ denote the normal components of the proximal ($p_l$) and distal ($d_l$) components of each limb, and $N = A \times B$ represents the normal vector for the plane defined by the hips and spine joints. In Fig.~\ref{fig:limbs_loss} we show a visual representation of $\mathcal{L}_{limbs}$ to better convey the intuitive reasoning behind its mathematical formulation.

\subsection{Pose Lifter Network (LiftNet)}
\label{sec:liftnet}

\begin{figure}
    \centering
    \includegraphics[width=0.96\textwidth]{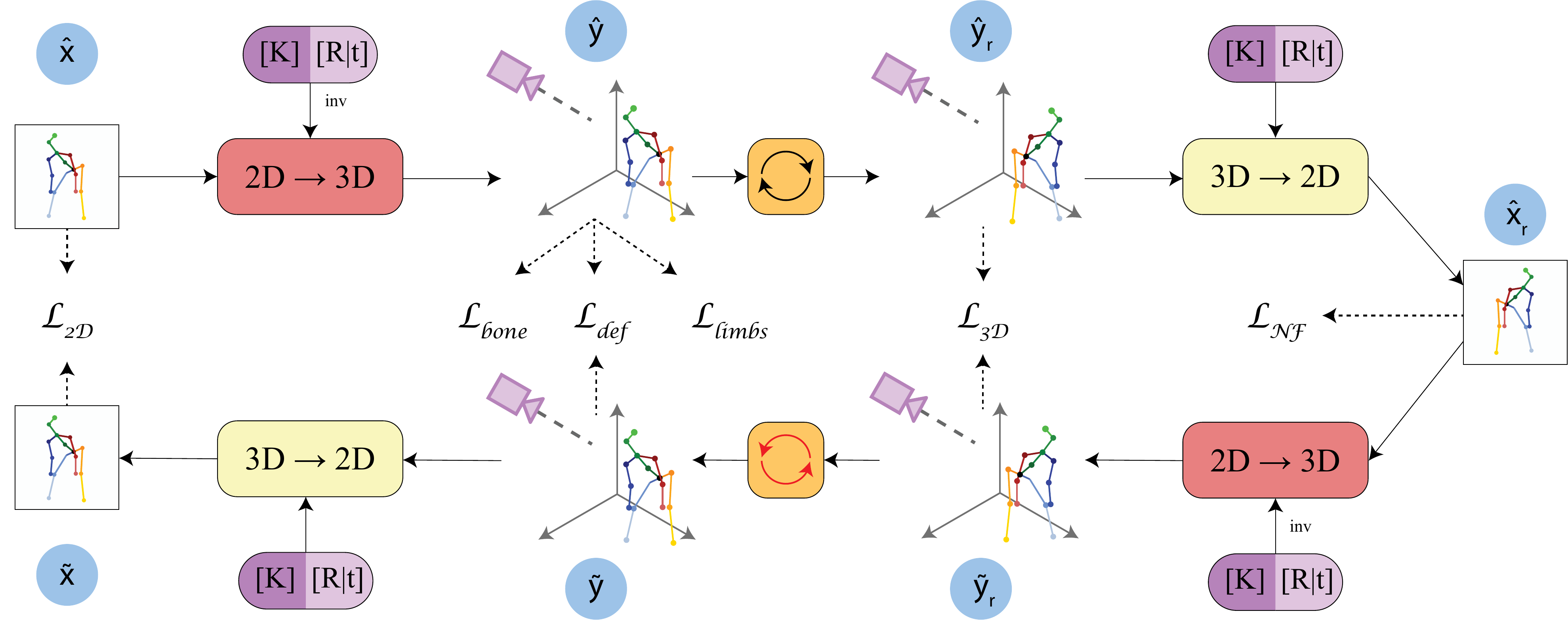}
    \caption{\textbf{LiftNet} architecture. The red ($2D \rightarrow 3D$), orange ($\circlearrowleft$ and $\circlearrowright$) and yellow ($3D \rightarrow 2D$) blocks describe the \textit{Lift}, \textit{Rotate}, \textit{Project} operations respectively. The symbol $x$ denotes a 2D pose, $y$ denotes a 3D pose. The decorator $\,\hat{}\,$ symbolizes a prediction in the forward pass while $\,\widetilde{}\,$ marks a prediction in the backward pass. The subscript $_r$ stands for rotated. The solid arrows describe the flow of the network, while the dashed arrows connect each intermediate datum to its loss.}
    \label{fig:pose_refiner}
\end{figure}

\textbf{LiftNet} is our lifter module which introduces the paradigm shown in Fig.~\ref{fig:teaser}(d). The overall detailed architecture is shown in Fig.~\ref{fig:pose_refiner}.

\textbf{LiftNet} aims at retrieving the 3D pose $y \in \mathbb{R}^{3J}$, starting from a 2D pose $\hat{x} \in \mathbb{R}^{2J}$ and its estimated camera parameters $[K]$ and $[R|T]$.
As shown in Fig.~\ref{fig:pose_refiner}, our architecture consists of a cycle consistency structure which can be split into two symmetric branches: a forward branch ($\hat{x} \rightarrow Lift \rightarrow \hat{y} \rightarrow Rotate \rightarrow \hat{y}_r \rightarrow Project \rightarrow \hat{x}_r$) and a backward branch ($\hat{x}_r \rightarrow Lift \rightarrow \widetilde{y}_r \rightarrow Inverse Rotate \rightarrow \widetilde{y} \rightarrow Project \rightarrow \widetilde{x}$). Each step and its input/output are described in Alg.~\ref{alg:cycle_consistency}. The lift operation is performed by a lifter network while the projection is a mathematical operation. Both of these operations rely on the full perspective model using camera parameters $[K][R|t]$. All the losses provide self-supervision to the cycle consistency, which does not access either 2D or 3D ground truths, making it a fully unsupervised process.

\begin{algorithm}
    \caption{Cycle consistency. Both $2D\rightarrow3D$ and $3D\rightarrow2D$ rely on the camera parameters $[K][R|t]$ to be solved.}\label{alg:cycle_consistency}
    \begin{algorithmic}
        \Require $\hat{x}$, $[K][R|t]$

      \State \textbf{Forward branch}  
        \State  1. \textit{Lift}: $2D\rightarrow3D$: $\hat{x} \rightarrow \hat{y}$
        \State  2. \textit{Rotate}: $\hat{y} \rightarrow \hat{y}_r$
        \State  3. \textit{Project}: $3D\rightarrow2D$: $\hat{y}_r \rightarrow \hat{x}_r$
      \State \textbf{Backward branch}
        \State  4. \textit{Lift}: $2D\rightarrow3D$: $\hat{x}_r \rightarrow \Tilde{y}_r$
        \State  5. \textit{InverseRotate}: $\Tilde{y}_r \rightarrow \Tilde{y}$
        \State  6. \textit{Project}: $3D\rightarrow2D$: $\Tilde{y} \rightarrow \Tilde{x}$
    \end{algorithmic}
\end{algorithm}

Differently from previous approaches using the weak camera model \cite{wandt2022elepose}, our lifter leverages the full perspective camera model to recover the 3rd dimension $w$ for each input 2D pose $\hat{x}$. Using the estimated $w$ we can solve the inverse of the projection of (Eq.~\ref{eq:full_camera_reprojection}) and recover the 3D joint positions. This \textit{Lift} operation is symbolized as $2D\rightarrow3D$.

Given a 3D poses $\hat{y}$ we can perform the \textit{Project} operation symbolized as $3D\rightarrow2D$. That means computing the 2D pose $\hat{x}$ using the full camera projection in Eq.~\ref{eq:full_camera_reprojection}. 

Inspired by previous approaches \cite{martinez2017simple,wandt2022elepose}, our lifter network consists of a simple MLP structure. The MLP receives as input a 2D pose $x$ concatenated with the flattened version the extrinsic parameters $[R|t]$ (12 total values), the intrinsic parameters $f=(f_w,f_h)$, $c=(c_w,c_h)$ and the scaling factor $s=(s_w,s_h)$, resulting in a vector of size $2J+18$. The input vector is fed (I) to a linear layer to obtain an embedded vector of size $dim_l$, (II) to 3 residual blocks each containing 2 fully connected layers, (III) to a linear layer to obtain the output vector of size $J$ representing the depth parameter $w$ for each joint. The vector is concatenated with the input $\hat{x}$ resulting in the estimated 3D pose $\hat{y}$.



To train the \textbf{LiftNet}, we minimize the following loss:
\begin{equation}
    \begin{aligned}
            \mathcal{L}_{lift} = \mathcal{L}_{2D}(\hat{x},\Tilde{x}) &+ \mathcal{L}_{3D}(\hat{y}_r,\Tilde{y}_r) + \mathcal{L}_{NF}(\Tilde{x}_r) + 
             \mathcal{L}_{bone}(\hat{y}) + \mathcal{L}_{limbs}(\hat{y}) + \mathcal{L}_{def}
    \end{aligned}
\end{equation}
where:
\begin{itemize}
    \item $\mathcal{L}_{2D}(\hat{x},\Tilde{x}) = ||\hat{x}-\Tilde{x}||_1$ is the norm-1 distance between the initial prediction $\hat{x}$ and its version after the cycle consistency loop $\Tilde{x}$;
    \item $\mathcal{L}_{3D}(\hat{y}_r,\Tilde{y}_r) = ||\hat{y}-\Tilde{y}||_2$ is the norm-2 distance between the 3D pose $\hat{y}$ and its version after it has been projected and lifted $\Tilde{y}$;
    \item $\mathcal{L}_{NF}(\Tilde{x}_r)$, $\mathcal{L}_{bone}(\hat{y})$ and $\mathcal{L}_{limbs}(\hat{y})$ are the losses defined in Sec.~\ref{sec:preliminaries};
    \item $\mathcal{L}_{def}$ is a deformation loss computed between two poses $y_a$ and $y_b$ belonging to the same batch defined as
    \begin{equation}
    \mathcal{L}_{def} = ||(\hat{y_a}-\hat{y_b}) - (\Tilde{y_a}-\Tilde{y_b})||_2
    \end{equation}
    which ensures that 2 poses from the same batch have not been deformed in completely different manners by the same \textit{Project} and \textit{Lift} operations, providing a supervision similar to the temporal consistency defined in \cite{yu2021towards}, but without relying on temporally-related data.
\end{itemize}







\begin{figure}
    \centering
    \includegraphics[width=0.96\textwidth]{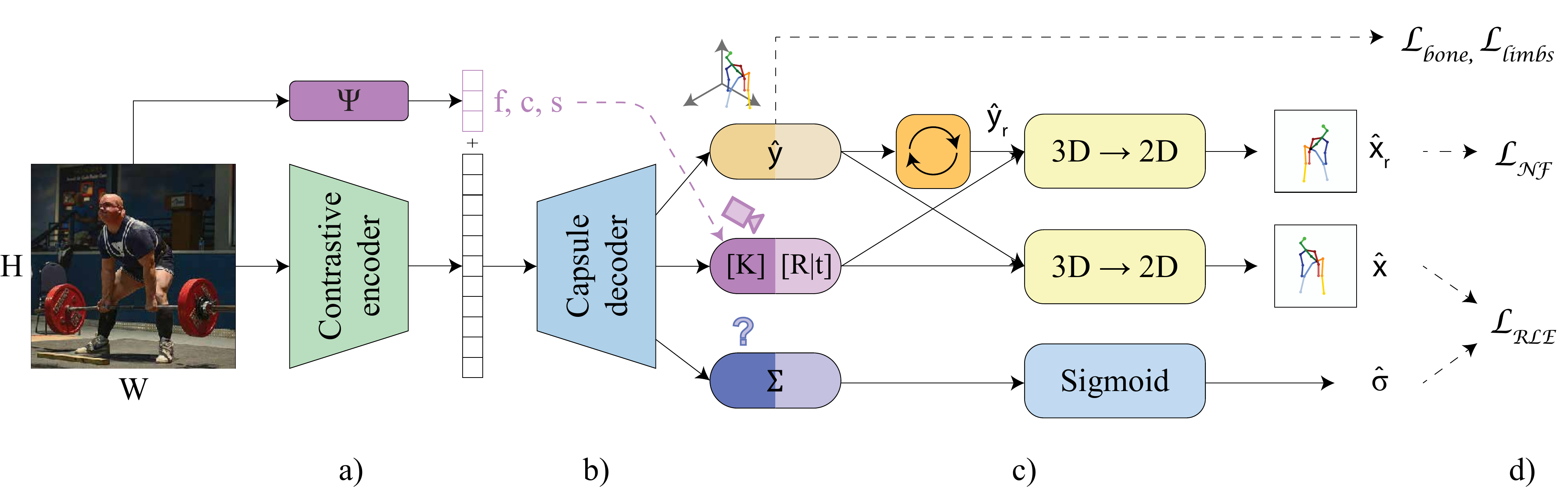}
    \caption{\textbf{RegNet} architecture. The $W \times H$ input image is fed to (a) a contrastive-pretrained encoder and a separate module $\Psi$ that estimates the intrinsic parameters. The output features are then concatenated and (b) fed into our attention-based capsule decoder. The outputs are three separate capsule vectors, representing an estimation of the 3D pose $\hat{y}$, of the camera $[K] [R|t]$ and a joint presence vector $\Sigma$. (c) Each of the outputs needs to be further processed before the loss computation. A copy of $\hat{y}$ is randomly rotated around the vertical axis, obtaining $\hat{y}_r$. $\hat{y}$ and $\hat{y}_r$ are projected into the camera plane and $\Sigma$ goes through a sigmoid activation function. (d) $\hat{y}$, $\hat{x}_r$, $\hat{x}$ and $\hat{\sigma}$ are fed to the loss functions.}
    \label{fig:pose_regressor}
\end{figure}

\subsection{Pose Regressor Network (RegNet)}
\label{sec:regnet}

RegNet is our direct regression module (Fig.~\ref{fig:teaser}(c)) which is used to estimate the camera pose and initial 2D pose used for the lifting stage. The overall detailed architecture is shown in Fig.~\ref{fig:pose_regressor}.
The input to \textbf{RegNet} is a single square image $I$ of size $W \times H$ pixels, roughly centered on the pelvis similar to \cite{wandt2022elepose,chen2019unsupervised,yu2021towards}. The objective is to retrieve the 2D pose $x \in \mathbb{R}^{3J}$. Additionally, it estimates the intrinsic camera parameters $K$ consisting of focal length $f = (f_w, f_h)$, principal point $c = (c_w,c_h)$, and a scaling factor $s = (s_w, s_h)$, as well as the extrinsic camera parameters $R$ and $t$. The intrinsic camera parameters are estimated directly from the image size as shown in \cite{li2022cliff}. 
 
RegNet consists of an encoder-decoder architecture, where the encoder is pre-trained using contrastive learning and the decoder is based on capsules. The output of the decoder yields both 3D poses and camera parameters, which are then used to compute the 2D pose $x \in \mathbb{R}^{2J}$, with each joint represented as $[u/w, v/w]$ on the image plane $I$. It is worth noting that we employ 2D poses for the loss computation, thus introducing a form of weak supervision to the 3D pose estimation process. 



\textbf{Contrastive Encoder.}
Our encoder employs a ResNet50 pre-trained using contrastive learning on ImageNet as in \cite{caron2020unsupervised}. Despite its general image focus, the use of contrastive learning allows for faster convergence and better generalization, outperforming supervised pre-training methods \cite{chen2020simclr}.

The output vector of the encoder is concatenated with the intrinsic parameters $[K]$ that are estimated directly from the full image. The vector resulting from the concatenation of size $dim$ is given as input to the decoder.

\textbf{Capsule-based decoder.} In \cite{kosiorek2019stacked}, they first showed how a Soft Attention mechanism can be effectively used to split a feature vector in different capsule features. In \cite{sabour2021unsupervised}, they perform an equivalent operation with a fully connected and a Softmax layer. Inspired by \cite{sabour2021unsupervised}, we design our capsule-based decoder using a Conv2D fully connected layer to transform the latent space vector of size $dim$ to a vector of size $9\times J$. 


Given $J$ values used for the attention mechanism, the remaining $8\times J$ values are divided in the following capsules:
\begin{itemize}
    \item $\hat{y} \in \mathbb{R}^{3J}$, representing the estimated 3D pose;
    \item $\Gamma \in \mathbb{R}^{3J}$ from which we can compute $[R|t]$ representing the extrinsic parameters, namely the estimated rotation matrix $R$ and translation vector $t$ with respect to the input image's viewpoint. The mathematical calculations to derive $[R|t]$ from $\Gamma$ are reported in the Supplementary Materials. To ensure invertibility, an orthogonality constraint is enforced on the rotation matrix $[R]$. The extrinsic parameters are combined with the intrinsic camera matrix $[K]$ computed by $\Psi$ at the encoding stage to obtain the full perspective camera model descriptor $[K][R|t]$;
    \item $\Sigma \in \mathbb{R}^{2J}$, a vector indicating the presence of each joint. Low values indicate uncertain detection of joints, often due to occlusions or joints being outside the image space.
\end{itemize}

\textbf{Outputs.} To ensure that the estimated 3D pose is plausible from multiple viewpoints, we require both the 2D pose $\hat{x}$ of the original image $I$, as well as the 2D pose $\hat{x}_r$ corresponding to the same pose from a different viewpoint. 

For $\hat{x}$, we perform the \textit{Project} operation as used in the lifting architecture to obtain 2D pose $\hat{x}$ using the full camera projection in Eq.~\ref{eq:full_camera_reprojection} given the 3D pose $\hat{y}$. 

Similar to the lifting stage, we observed that we can increase the accuracy of the estimated 3D poses by using a Normalizing Flow loss to ensure the plausibility of multiple 2D projections of the 3D pose seen from different viewpoints. To this end we also calculate a rotated projection $\hat{x}_r$, where we rotate the model around the vertical world axis. The \textit{Rotate} operation is applied randomly to each 3D pose in the range of $[\ang{10},\ang{350}]$ to simulate a viewpoint change, before projecting the rotated 3D pose $\hat{y}_r$ to obtain $\hat{x}_r$.

Regression of the joints' position can be combined with an uncertainty measure and leads to better results when compared to direct regression of joints and heatmap-based methods~\cite{li2021human}. Moreover, the regression is more computation and memory efficient compared to heatmaps. To employ a similar method, we estimate the deviation $\hat{\sigma}$ of the predicted joint's position from the ground truth by applying a sigmoid function on the estimated presence capsule $\Sigma$.




\textbf{Losses.} To train \textbf{RegNet}, we need to minimize the following loss:
\begin{equation}
    \begin{aligned}
    \mathcal{L}_{reg} = \mathcal{L}_{bone}(\hat{y}) + \mathcal{L}_{limbs}(\hat{y}) &+ \mathcal{L}_{NF}(\hat{x}_r) + \mathcal{L}_{RLE}(\hat{x}, \hat{\sigma})
    \end{aligned}
\end{equation}
where:
\begin{itemize}
    \item $\mathcal{L}_{NF}(\Tilde{x}_r)$, $\mathcal{L}_{bone}(\hat{y})$ and $\mathcal{L}_{limbs}(\hat{y})$ are the losses defined in Sec.~\ref{sec:preliminaries};
    \item $\mathcal{L}_{RLE}(x,I)$ is the residual log-likelihood estimation loss defined as,
    \begin{equation}
    \begin{aligned}
        \mathcal{L}_{RLE}(x,I) &= -log P_{\Theta,\phi}(x,I)|_{x=\hat{x}}
        = -log P_{\phi}(\hat{x}) + log (\hat{\sigma})
    \end{aligned}
    \end{equation}
    As in \cite{li2021human}, this loss aims at estimating the joints' position $\hat{x}$ via direct regression coupled with the learned error distribution $\hat{\sigma}$. We refer to the original work \cite{li2021human} for the full mathematical explanation of the loss.
\end{itemize}

For further details on loss balancing during training see the Supplementary Materials.





\section{Results}
\label{sec:results}

We perform our experiments on the common Human3.6M \cite{ionescu2013human3} and MPI-INF-3DHP \cite{mono-3dhp2017} datasets\footnote{All datasets were obtained and used only by the authors affiliated with academic institutions.}
.
We follow standard test protocols for both datasets \cite{wandt2022elepose}. We also report extensive ablation studies and qualitative results on the unseen 3DPW dataset \cite{von2018recovering} to demonstrate generalization.

\textbf{Implementation details.} The input images are of size $H=224$ and $W=224$. The skeleton model has $J=17$ joints.
\textbf{RegNet} is trained for $45$ epochs using the optimizer AdamW \cite{loshchilov2018fixing}, $dim = 2048 + 6$, learning rate $1e-3$, and weight decay $1e-4$.
\textbf{LiftNet} is trained for $100$ epochs using the optimizer AdamW, learning rate $2e-4$, and weight decay $1e-5$. 
Both \textbf{RegNet} and \textbf{LiftNet} are trained with batch size $256$ and bfloat16 precision on single a NVIDIA RTX 3090. Inference runs on the same GPU at $\approx45$ fps.

\textbf{Metrics.} We adopt the standard mean per joint position error (MPJPE) in two common forms for the 3D HPE unsupervised settings: the PA-MPJPE where reconstructed 3D pose is Procrustes aligned and the N-MPJPE where the 3D pose is normalized the same scale as the ground truth \cite{rhodin2018unsupervised}. As in \cite{wandt2022elepose}, for the MPI-INF-3DHP dataset we report the scale normalized percentage of correct key points (N-PCK) predicted within $150$ mm to the original position and its corresponding area under curve (AUC).

\subsection{Quantitative results}

\begin{table}[t]
\centering
\begin{minipage}[t]{.49\textwidth}
  \centering
  \resizebox{\textwidth}{!}{%
    \begin{tabular}{@{}ccc@{}c}
    \toprule
    \textbf{Supervision}& \textbf{Method}                 & \textbf{PA-MPJPE}$\downarrow$ &\textbf{N-MPJPE} $\downarrow$\\ \midrule
     Full-3D& Pavalkos \cite{pavlakos2018ordinal}& 41.8&-\\ \midrule
    \multirow{3}{*}{10\%-3D}& Kundu \cite{kundu2022uncertainty}& 49.6&59.4\\
     & Kundu \cite{kundu2021non}& 48.2&57.6\\
     & Gong \cite{gong2021poseaug}& \textbf{39.1}&\textbf{50.2}\\ \midrule
    \multirow{3}{*}{Multi-view}& Rhodin \cite{rhodin2018unsupervised}    & 98.2           &122.6\\
                                    & Kundu \cite{kundu2020self}              & 85.8           &-\\ 
                                    & \textbf{Usman} \cite{usman2022metapose} & \textbf{44.0}&\textbf{55.0}\\ \midrule
                                    \multirow{3}{*}{Full-2D}& Fish \cite{fish2017adversarial}         & 97.2           &-\\
                                    & Kundu \cite{kundu2020self}                 & 62.4           &-\\
                                    & \textbf{Ours (RegNet)}& \textbf{45.4}  &\textbf{69.9}\\ \bottomrule
    \end{tabular}%
  }
  \caption{Quantitative results for \textbf{direct regression from images} on Human3.6M.}
  \label{tab:regression_results}
\end{minipage}\hfill
\begin{minipage}[t]{.45\textwidth}
  \centering
  \resizebox{\textwidth}{!}{%
    \begin{tabular}{@{}cccc@{}}
    \toprule
    \textbf{Supervision}                   & \textbf{Method}                 & \textbf{PA-MPJPE}$\downarrow$      & \textbf{N-MPJPE}$\downarrow$       \\ \midrule
    Full                          & Martinez \cite{martinez2017simple}              & 37.1          & 45.5         \\ \midrule
    \multirow{4}{*}{Weak}         & Fish \cite{fish2017adversarial}                   & 79.0          & -             \\
                                  & Wandt \cite{wandt2019repnet}                 & 38.2          & 50.9          \\
                                  & Drover \cite{drover2018can}                & 38.2          & -             \\
                                  & Kundu \cite{kundu2022uncertainty}                 & 62.4          & -             \\ \midrule
    \multirow{2}{*}{Multi-view}   & Kocabas \cite{kocabas2019self}          & 47.9          & 54.9          \\
                                  & Wandt \cite{wandt2021canonpose}                 & 51.4          & 65.9          \\ \midrule
    \multirow{4}{*}{Unsupervised} & Chen \cite{chen2019unsupervised}                  & 58.0          & -             \\
                                  & Yu \cite{yu2021towards}         & 42.0          & 85.3         \\ 
                                  & Wandt \cite{wandt2022elepose}                 & 36.7          & 64.0          \\
                                  & \textbf{Ours (LiftNet)} & \textbf{30.7} & \textbf{50.8} \\ \bottomrule
    \end{tabular}%
  }
  \caption{Quantitative results for \textbf{lifting from 2D ground truth} on Human3.6M.}
  \label{tab:quantitative_gt}
\end{minipage}
\end{table}

\begin{table}[t]
\centering
\begin{minipage}[t]{.49\textwidth}
  \centering
\resizebox{\textwidth}{!}{%
\begin{tabular}{@{}cccc@{}}
\toprule
\textbf{Method} &\textbf{Backbone}& \textbf{PA-MPJPE$\downarrow$} & \textbf{N-MPJPE$\downarrow$} \\ \midrule
Chen \cite{chen2019unsupervised}&SH \cite{newell2016stacked}& 68.0                          & -                            \\
 Wandt \cite{wandt2019repnet}& SH \cite{newell2016stacked}& 65.1&89.9\\
 Kundu \cite{kundu2020self}& ResNet-50& 63.8                          &-                            \\
 Kundu \cite{kundu2020self}& ResNet-50& 62.4                          &-                            \\
Yu \cite{yu2021towards}&CPN \cite{chen2018cascaded}& 52.3                          & 92.4                        \\
Wandt \cite{wandt2022elepose}&CPN \cite{chen2018cascaded}& 50.2                          & 74.4                         \\
\textbf{Ours (LiftNet)}  & \textbf{Ours (RegNet)}& \textbf{43.8}                 & \textbf{67.1}                \\ \bottomrule
\end{tabular}
}
\caption{Quantitative results for \textbf{lifting from 2D predictions} on Human3.6M.}
\label{tab:quant_results}
\end{minipage}\hfill
\begin{minipage}[t]{.45\textwidth}
  \centering
\resizebox{\textwidth}{!}{%
\begin{tabular}{@{}cccccc@{}}
\toprule
\textbf{Supervision}                   & \textbf{Method}                &\textbf{2D Input}& \textbf{PA-MPJPE}$\downarrow$& \textbf{N-PCK}$\uparrow$& \textbf{AUC}$\uparrow$\\ \midrule
Weak                          & Kundu  \cite{kundu2020self}                 &GT& 93.9           & 84.6  & 60.8  \\ \midrule
\multirow{4}{*}{Unsup.} & Yu \cite{yu2021towards}                   &GT& -              & 86.2 & 51.7 \\
                              & Wandt \cite{wandt2022elepose}                &GT& 54.0           & 86.0  & 50.1  \\
                              & \textbf{Ours (LiftNet)}&\textbf{Pred}& \textbf{46.8} & \textbf{93.6}     & \textbf{61.3}     \\
                              & \textbf{Ours (LiftNet)}&\textbf{GT}& \textbf{33.6} & \textbf{97.6}     & \textbf{69.5}     \\
                               \bottomrule

Unseen data                      & \textbf{Ours (LiftNet)}&\textbf{Pred}& \textbf{57.3} & \textbf{77.5}     & \textbf{46.4}     \\ \bottomrule
\end{tabular}
}
\caption{Quantitative results for 3D HPE on MPI-INF-3DHP.}
\label{tab:mpi_results}
\end{minipage}
\end{table}

\textbf{Direct regression from images.} In Tab.~\ref{tab:regression_results}, we report the results of weakly-supervised direct regression of 3D pose from images on Human3.6M dataset. Weakly-supervised approaches include ones using only a small portion of 3D annotated data (10\%-3D) or multi-view supervision. Both approaches rely on 3D spatial information for their supervision, leading to results close to the baseline fully supervised approach \cite{pavlakos2018ordinal}. Among approaches using only 2D information, our \textbf{RegNet} outperforms others, obtaining results in line with baseline 3D supervised. In contrast to the previous approaches, \textbf{RegNet} also performs the unsupervised estimation of the full perspective camera parameters $[K][R|t]$.

\textbf{Lifting from 2D ground truth (GT).} In Tab.~\ref{tab:quantitative_gt}, we report the results of lifting GT 2D pose to 3D on Human3.6M dataset. We compare against baseline fully supervised approaches \cite{martinez2017simple}, multi-view supervised approaches \cite{kocabas2019self,wandt2021canonpose}, and weakly supervised approaches adopting different strategies like domain adaptation \cite{kundu2022uncertainty}, 2D GT poses \cite{drover2018can,fish2017adversarial} and partial 3D GT \cite{wandt2019repnet}. Our unsupervised \textbf{LiftNet} outperforms all previous weakly supervised approaches, also obtaining the best results among unsupervised methods. We also outperform \cite{wandt2022elepose} which adopts a weak perspective camera modeling combined with elevation estimation, demonstrating the efficacy of our \textbf{LiftNet} that explicitly leverages the full perspective camera model.

\textbf{Lifting from 2D predictions.} In Tab.~\ref{tab:quant_results}, we report the results of lifting 2D pose predictions to 3D on the Human3.6M dataset. In contrast to other unsupervised approaches that use either ResNet50 \cite{kundu2020self}, Stacked Hourglass (SH) \cite{chen2019unsupervised,wandt2019repnet} or Cascaded Pyramid Network (CPN) \cite{chen2018cascaded,wandt2022elepose} as the backbone to extract 2D predictions to be lifted, we use \textbf{RegNet} which also provides an unsupervised estimation of the camera parameters. The full EPOCH approach combining both \textbf{RegNet} and \textbf{LiftNet} outperforms all other methods, demonstrating its efficacy as an end-to-end approach to estimate 3D poses from images.


\textbf{Generalising to unseen data.} In Tab.~\ref{tab:mpi_results}, we report the results of 3D HPE on MPI-INF-3DHP dataset. When trained on MPI-INF-3DHP, \textbf{LiftNet} outperforms both weakly-supervised \cite{kundu2020self} and unsupervised \cite{wandt2022elepose,yu2021towards} lifting methods starting from either the ground truth or from poses predicted by \textbf{RegNet}. When not trained on MPI-INF-3DHP (last row), EPOCH trained only on Human3.6M achieves results comparable to fine-tuned approaches, proving its ability to generalize to unseen data.

\textbf{Ablation studies.} In Tab.~\ref{tab:regnet_ablations}, we report the results of the ablation study for \textbf{RegNet}. First, we show how using a backbone trained with supervised learning (S) leads to poorer features (first row), leading to performance degradation compared to the full model trained with contrastive learning (C) (last row). Next, we ablate different losses, showing how $\mathcal{L}_{NF}$ is the loss that causes the biggest drop in performances since it ensures the network does not estimate 3D poses that are plausible only from a single viewpoint. $\mathcal{L}_{bone}$ and $\mathcal{L}_{limbs}$ have similar effect on the results as they both enforce anthropomorphic constraints. 

In Tab.~\ref{tab:liftnet_ablations}, we report the results of the ablation study for \textbf{LiftNet}. We perform an ablation study on the camera modeling, showing how using a weak perspective camera leads to performance degradation due to its less precise modeling of the 2D/3D relation than the full perspective camera. Moreover, we proceed to study the effect of each loss on the performances. As for \textbf{RegNet}, $\mathcal{L}_{NF}$ causes the biggest drop in performances. The deformation $\mathcal{L}_{def}$, and the two anthropomorphic constraints $\mathcal{L}_{bone}$ and $\mathcal{L}_{limbs}$ have similar effects on the numerical results, as they are all enforcing comparable constraints on the deformation and proportions of 3D poses.
$\mathcal{L}_{2D}$ and $\mathcal{L}_{3D}$ all provide regularisation between different stages of the cycle consistency, so removing one of them has a comparable effect on the performance's degradation, since \textbf{LiftNet} is still regularised by the remaining ones.

\begin{table}[t]
\centering
\begin{minipage}[t]{.45\textwidth}
  \centering
  \resizebox{\textwidth}{!}{%
\begin{tabular}{@{}ccclcc@{}}
\toprule
Backbone                     & $\mathcal{L}_{NF}$                     & \begin{tabular}[c]{@{}c@{}}$\mathcal{L}_{bone}$\end{tabular}&\begin{tabular}[c]{@{}c@{}}$\mathcal{L}_{limbs}$\end{tabular}& PA-MPJPE$\downarrow$ & MPJPE$\downarrow$ \\ \midrule
S                            & {\color[HTML]{34FF34} \cmark}          & {\color[HTML]{34FF34} \cmark}                                                           &{\color[HTML]{34FF34} \cmark}                                                           &  60.1                 & 82.7              \\ \midrule
                             & {\color[HTML]{FE0000} \xmark}          & {\color[HTML]{34FF34} \cmark}                                                                  &{\color[HTML]{34FF34} \cmark}                                                                  & 239.2                & 342.8             \\
 & {\color[HTML]{34FF34} \cmark}& {\color[HTML]{FE0000} \xmark}& {\color[HTML]{34FF34} \cmark}                                                                  & 49.1&71.4\\
                             & {\color[HTML]{34FF34} \cmark}          & {\color[HTML]{34FF34} \cmark}&{\color[HTML]{FE0000} \xmark}                                                                   & 48.3& 70.9\\
\multirow{-4}{*}{\textbf{C}}& {\color[HTML]{34FF34} \textbf{\cmark}} & {\color[HTML]{34FF34} \textbf{\cmark}}                                                   &{\color[HTML]{34FF34} \textbf{\cmark}}                                                   & \textbf{45.4}        & \textbf{69.9}     \\ \bottomrule
\end{tabular}
}
\caption{Ablation study for \textbf{RegNet}. The last row corresponds to our full model. S = Supervised. C = Contrastive.}
\label{tab:regnet_ablations}
\end{minipage}\hfill
\begin{minipage}[t]{.45\textwidth}
  \centering
\resizebox{\textwidth}{!}{%
\begin{tabular}{@{}ccccccccc@{}}
\toprule
\begin{tabular}[c]{@{}c@{}}Camera\\ model\end{tabular}                         & $\mathcal{L}_{NF}$                     & \begin{tabular}[c]{@{}c@{}}$\mathcal{L}_{bone}$\end{tabular}& \begin{tabular}[c]{@{}c@{}}$\mathcal{L}_{def}$\end{tabular}&\begin{tabular}[c]{@{}c@{}}$\mathcal{L}_{limbs}$\end{tabular}& $\mathcal{L}_{2D}$                     & $\mathcal{L}_{3D}$                     & PA-MPJPE$\downarrow$      & MPJPE$\downarrow$         \\ \midrule
W                            & {\color[HTML]{34FF34} \cmark}          & {\color[HTML]{34FF34} \cmark}                                                                                   & {\color[HTML]{34FF34} \cmark}                                                                                   &{\color[HTML]{34FF34} \cmark}                                                                                   & {\color[HTML]{34FF34} \cmark}          & {\color[HTML]{34FF34} \cmark}          & 47.7          & 84.9          \\ \midrule
                                & {\color[HTML]{FE0000} \xmark}          & {\color[HTML]{34FF34} \cmark}                                                                                   & {\color[HTML]{34FF34} \cmark}                                                                                   &{\color[HTML]{34FF34} \cmark}                                                                                   & {\color[HTML]{34FF34} \cmark}          & {\color[HTML]{34FF34} \cmark}          & 232.6         & 293.0         \\
 & {\color[HTML]{34FF34} \cmark}          & {\color[HTML]{FE0000} \xmark}                                                                                   & {\color[HTML]{34FF34} \cmark}& {\color[HTML]{34FF34} \cmark}& {\color[HTML]{34FF34} \cmark}          & {\color[HTML]{34FF34} \cmark}          & 41.5&56.4\\
 & {\color[HTML]{34FF34} \cmark}          & {\color[HTML]{34FF34} \cmark}& {\color[HTML]{FE0000} \xmark}                                                                                   & {\color[HTML]{34FF34} \cmark}& {\color[HTML]{34FF34} \cmark}          & {\color[HTML]{34FF34} \cmark}          & 41.3&55.8\\
                                & {\color[HTML]{34FF34} \cmark}          & {\color[HTML]{34FF34} \cmark}& {\color[HTML]{34FF34} \cmark}&{\color[HTML]{FE0000} \xmark}                                                                                   & {\color[HTML]{34FF34} \cmark}          & {\color[HTML]{34FF34} \cmark}          & 40.9& 54.9\\
                                & {\color[HTML]{34FF34} \cmark}          & {\color[HTML]{34FF34} \cmark}                                                                                   & {\color[HTML]{34FF34} \cmark}                                                                                   &{\color[HTML]{34FF34} \cmark}                                                                                   & {\color[HTML]{FE0000} \xmark}          & {\color[HTML]{34FF34} \cmark}          & 40.1          & 54.3          \\
                                & {\color[HTML]{34FF34} \cmark}          & {\color[HTML]{34FF34} \cmark}                                                                                   & {\color[HTML]{34FF34} \cmark}                                                                                   &{\color[HTML]{34FF34} \cmark}                                                                                   & {\color[HTML]{34FF34} \cmark}          & {\color[HTML]{FE0000} \xmark}          & 39.5          & 53.4          \\
\multirow{-7}{*}{\textbf{F}}& {\color[HTML]{34FF34} \textbf{\cmark}} & {\color[HTML]{34FF34} \textbf{\cmark}}                                                                          & {\color[HTML]{34FF34} \textbf{\cmark}}                                                                          &{\color[HTML]{34FF34} \textbf{\cmark}}                                                                          & {\color[HTML]{34FF34} \textbf{\cmark}} & {\color[HTML]{34FF34} \textbf{\cmark}} & \textbf{30.7} & \textbf{50.8} \\ \bottomrule
\end{tabular}
}
\caption{Ablation study for \textbf{LiftNet}. The last row correspond to our complete model. W = Weak perspective camera. F = Full perspective camera. 
}
\label{tab:liftnet_ablations}
\end{minipage}
\end{table}


\begin{figure*}[t]
    \centering
    \includegraphics[width=0.96\textwidth]{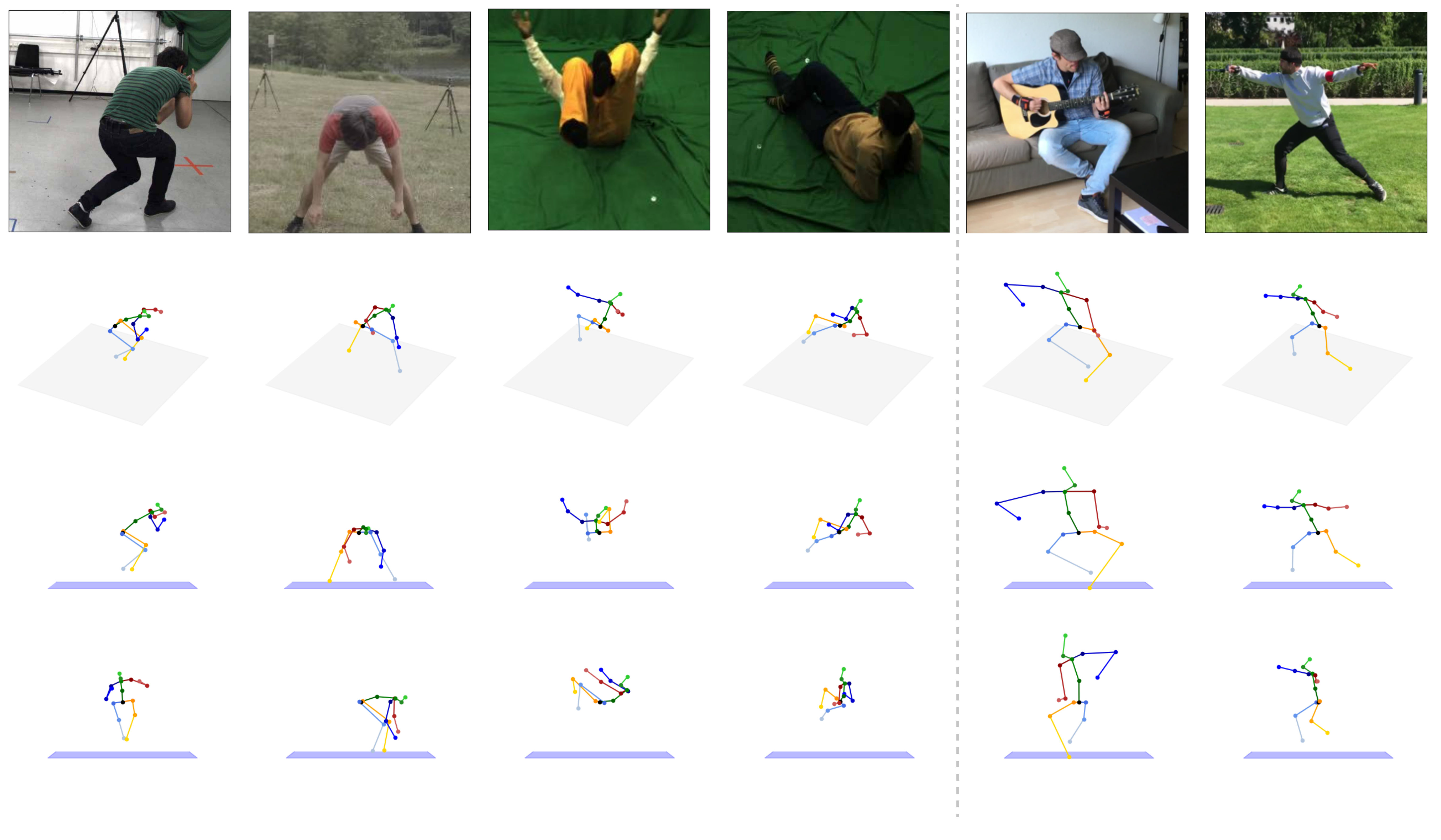}
    \captionsetup{skip=-5pt}
    \caption{EPOCH qualitative results on MPI-INF-3DHP~\cite{mono-3dhp2017} (columns 1, 2, 3, 4), 3DPW~\cite{von2018recovering} (columns 5, 6). Rows: input images, \textbf{RegNet} output, \textbf{LiftNet} output (front and side view). Our method can generalize to unseen in-the-wild data (3DPW) even if only trained on Human3.6M data.}
    \label{fig:qualitative}
\end{figure*}


\subsection{Qualitative results}

Fig.~\ref{fig:qualitative} shows our qualitative results on challenging poses for Human3.6M and MPI-INF-3DHP. Even in the presence of occlusions and rare poses (e.g. sitting on a chair, lying on the floor with crossed legs), both \textbf{RegNet} and \textbf{LiftNet} obtain visually plausible 3D poses. Moreover, we display results on 3DPW which is unseen at training time. Even if the scale of the 3D pose is different, we still obtain plausible poses from challenging images, demonstrating the ability of our EPOCH approach to generalize to unseen scenarios.

\section{Conclusions}
\label{sec:conclusions}

In this paper, we presented EPOCH, a novel framework that jointly estimates the 3D pose of cameras and humans consisting of \textbf{LiftNet} and \textbf{RegNet}. 
\textbf{LiftNet} performs the unsupervised 3D lifting starting from estimations of both 2D poses and camera parameters. To address the unavailability of camera parameters in real world scenarios, we design \textbf{RegNet}, a novel human pose regressor that can jointly estimate 2D and 3D poses as well as perspective camera parameters using weak 2D pose supervision.
We show that an estimated full perspective camera allows us to substantially improve the unsupervised 3D human pose estimation accuracy and consistency over state-of-the-art results. By estimating the camera only from 2D poses without any 3D or camera ground truth, we can generalise to unseen data, making a step forward towards fully unsupervised 3D HPE in-the-wild.

\clearpage
\setcounter{page}{1}
\title{Supplementary Materials } 
\maketitle


\section{Code}

The code will be made available in a Github repository upon acceptance.

\section{Details about the Normalizing Flows' invertible function}

As in \cite{wandt2022elepose}, we design a NF to estimate the plausibility of 2D poses. We train the NF to map each pose $x \in \mathbb{R}^{2J}$ onto a sample point $z$ of a Gaussian Distribution $p_z(z)$.
Given a 2D pose $x$ and an invertible function $f$ such that $\mathbf{f}(z) = x$, the PDF of $x$ can be computed as

\begin{equation}
    p_x(\mathbf{x}) = p_z(\mathbf{f}^{-1}(x)) \left| \det \left(\frac{\partial \mathbf{f}^{-1}}{\partial \mathbf{x}}\right) \right|,
\end{equation}

where $\frac{\partial \mathbf{f}^{-1}}{\partial \mathbf{x}}$ is the Jacobian matrix of the inverse transformation. The NF is trained offline to learn the PDF $p_z(\mathbf{f}^{-1}(x))$ of the 2D poses in a chosen training dataset. Once trained, the probability $p_x(\mathbf{x})$ can be computed for a new sample $x$ and used as a measure of its plausibility to lie within the
distribution of the learned dataset.



\section{Camera matrix notation}

In the paper we refer to the intrinsic $[K]$ and extrinsic $[R|t]$ camera matrices as detailed below.

The intrinsic matrix, representing the internal parameters of a camera, is defined as:
\[ [K] = \begin{bmatrix}
f_w & sk & c_w \\
0 & f_h & c_h \\
0 & 0 & 1
\end{bmatrix} \]
Where:
\begin{itemize}
    \item \(f_w\) and \(f_h\) are the focal lengths in the x and y directions.
    \item  \(sk\) represents skew, which is usually zero in most cameras. In the paper we assume $sk = 0$ for all cameras.
    \item \((c_w, c_h)\) is the principal point, the optical center of the image.
\end{itemize}

The extrinsic matrix, denoting the rotation and translation of the camera with respect to the world coordinate system, can be expressed as:
\[ [R | t] = \left[\begin{array}{ccc|c}
r_{11} & r_{12} & r_{13} & t_X \\
r_{21} & r_{22} & r_{23} & t_Y \\
r_{31} & r_{32} & r_{33} & t_Z
\end{array}\right] \]
Where:
\begin{itemize}
    \item $[R]$ represents the rotation matrix.
    \item \(t\) represents the translation vector.
\end{itemize}

\section{Mathematical derivation of the intrinsic parameters from the image}

The scaling parameters $s_w$ and $s_h$ in the paper are computed as follows:

\[ s_w = \frac{W}{W_{BB}*\mu_h}\]
\[ s_h = \frac{H}{H_{BB}*\mu_h}\]

where $\mu_h$ is the mean length of the vector from the root joint to the head joint, as in \cite{wandt2022elepose}. $W_{BB}, H_{BB}$ are the width and height of the bounding box in pixel prior scaling it to $224 \times 224$.

Taking inspiration from \cite{li2022cliff}, we estimate the focal length starting from the uncropped input image width $W_{full}$ and height $H_{full}$ as follows:

\[ f = \sqrt{W_{full}^2 + H_{full}^2}\]

and scale it along each axis

\[ f_w = f * s_w \]
\[ f_h = f * s_h \]

The principal point is computed as:

\[c_w = \Big(C_w - LEFT - \frac{W}{2}\Big) * s_w\]
\[c_h = \Big(C_h - TOP - \frac{H}{2}\Big) * s_h\]

where $C_w = \frac{W_{full}}{2}$, $C_h = \frac{H_{full}}{2}$. $LEFT$ and $TOP$ are the pixel coordinates of the top-left corner of the unscaled bounding box in the full-size image.

\section{Mathematical derivation of the extrinsic parameters from the capsule}

When predicting capsules, we estimate $\Gamma \in \mathbb{R}^{3J}$ from which we can compute the rotation matrix $[R]$ and translation vector $t$ with respect to the input image's viewpoint.

Starting from $\Gamma$ we compute the average vector $\bar{\Gamma} \in \mathbb{R}^{3}$. $\bar{\Gamma}$ is expressed as $[\theta_X, \theta_Y, w_p]$, where $\theta_X$ and $\theta_Y$ are the euler rotation angles in the $X$ and $Y$ world-space axis and $w_p$ is the distance from the pelvis to the camera along the Z axis. We discard the rotation along the $Z$ axis because that would lead to infinite configurations of the predicted camera and 3D pose. Thus, we set $\theta_Z=0$.

Starting from $[\theta_X, \theta_Y, 0]$, we can derive the rotation matrix as:

\begin{align*}
    [R] = 
    \left[\begin{array}{ccc}
    1 & 0 & 0 \\
    0 & \cos(\theta_X) & -\sin(\theta_X) \\ 
    0 & \sin(\theta_X) & \cos(\theta_X)
    \end{array}\right] & \left[\begin{array}{ccc}
    \cos(\theta_Y) & 0 & \sin(\theta_Y) \\
    0 & 1 & 0 \\
    -\sin(\theta_Y) & 0 & \cos(\theta_Y)
    \end{array}\right]
\end{align*}

Starting from $w_p$, we can derive the translation vector. We define a setting in which the pelvis is centered in $(X,Y,Z) = (0,0,0)$ in world coordinates and in $(u,v) = (0,0)$ on the image plane. Thus we can express the relation between the pelvis in 2D and 3D as:

\[\begin{bmatrix}
0 \\
0 \\
w_p
\end{bmatrix} =
\begin{bmatrix}
f_w & 0 & c_w \\
0 & f_h & c_h \\
0 & 0 & 1
\end{bmatrix}
\left[\begin{array}{ccc|c}
r_{11} & r_{12} & r_{13} & t_X \\
r_{21} & r_{22} & r_{23} & t_Y \\
r_{31} & r_{32} & r_{33} & t_Z
\end{array}\right]
\begin{bmatrix}
0 \\
0 \\
0 \\
1
\end{bmatrix}\]

from which we can obtain

\[\begin{bmatrix}
0 \\
0 \\
w_p
\end{bmatrix} =
\begin{bmatrix}
f_w*t_X + c_w*t_Z \\
f_h*t_Y + c_h*t_Z \\
t_Z
\end{bmatrix}
\]

thus deriving

\[\begin{bmatrix}
t_X \\
t_Y \\
t_Z
\end{bmatrix} =
\begin{bmatrix}
- \frac{c_w*w_p}{f_w} \\
- \frac{c_h*w_p}{f_h} \\
w_p
\end{bmatrix}
\]




\section{Optimization of Loss Function Weights}
In \textbf{RegNet} and \textbf{LiftNet}, we encounter a discrepancy in the scale of different components of the loss function. Specifically, certain losses such as $\mathcal{L}_{RLE}$ and $\mathcal{L}_{NF}$ are formulated as negative-log-likelihoods, resulting in predominantly highly negative values. In contrast, other components of the loss function yield positive values close to zero. To facilitate a more stable and efficient optimization process, we implement a scaling mechanism for $\mathcal{L}_{RLE}$ and $\mathcal{L}_{NF}$ to align their value ranges with the other loss components. More in detail, we shift the loss function vertically by its expected lowest value and divide it by its expected vertical span, to make the scaled loss fit the interval $[0,1]$.

Additionally, we introduce a non-trainable hyperparameter $\lambda$, set to $10$, to strategically weight certain loss components more heavily in the overall loss function. This approach allows for greater emphasis on key aspects of the learning objective.

For \textbf{RegNet}, we apply this weighting strategy to $\mathcal{L}_{RLE}$, recognizing its critical role in the network's performance. In the case of \textbf{LiftNet}, the weights are applied to $\mathcal{L}_{2D}$ and $\mathcal{L}_{bone}$. This ensures the integrity of the pose cycle consistency and maintains the correct proportions of the 3D skeletal model in terms of bone lengths.

In both \textbf{RegNet} and \textbf{LiftNet}, $\mathcal{L}_{limbs}$ is treated as a regularization term. Accordingly, it is scaled down by a factor of $0.1$, reflecting its role in preventing overfitting and ensuring generalization, rather than directly driving the primary learning objective.

\section{Additional qualitative results}

\begin{figure*}
    \centering
    \includegraphics[width=\textwidth]{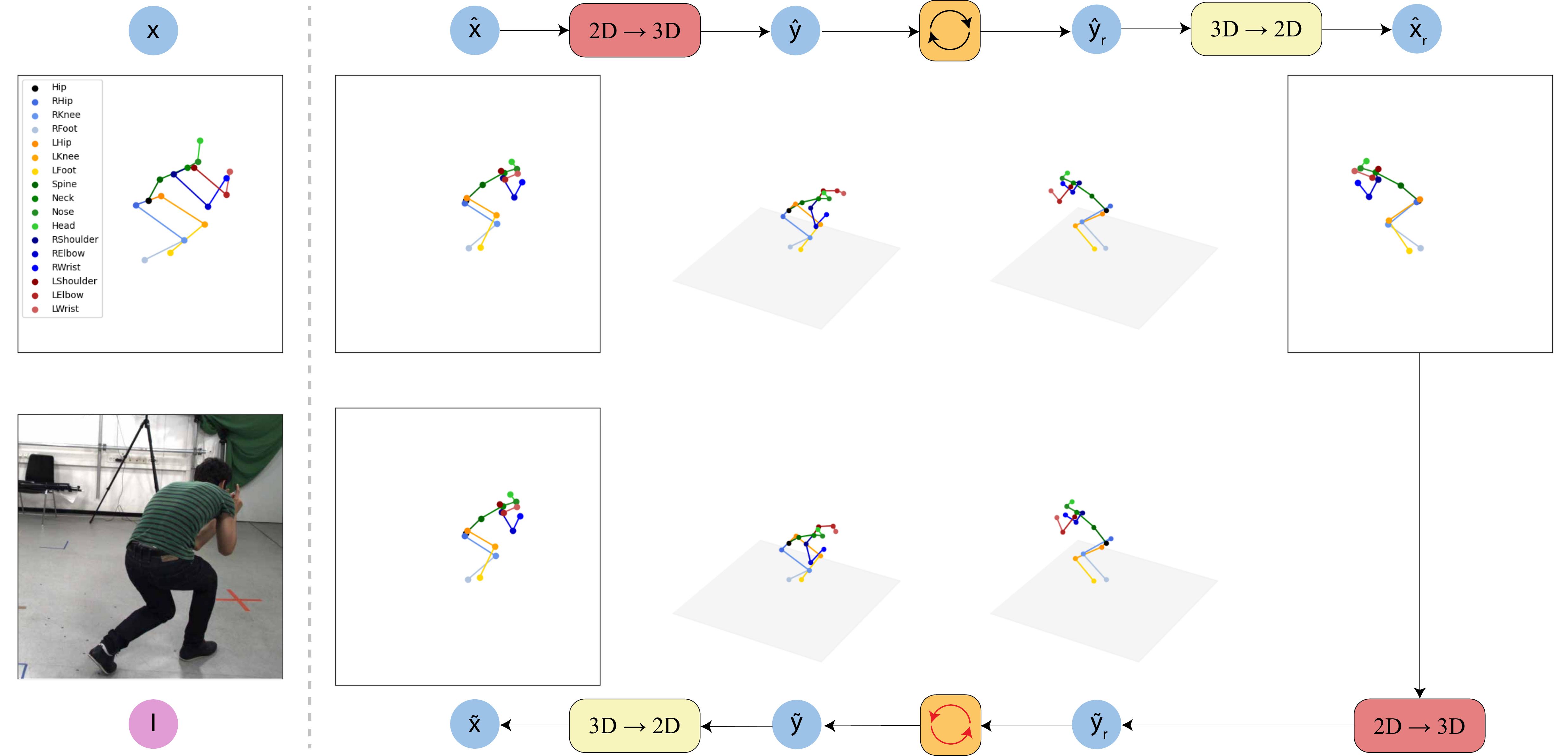}
    \caption{Intermediate results of our \textbf{LiftNet} architecture on MPI-INF-3DHP \cite{mono-3dhp2017}, as detailed in Fig.~\ref{fig:pose_refiner} in the main paper.}
    \label{fig:supp_res_1}
\end{figure*}

\begin{figure*}
    \centering
    \includegraphics[width=\textwidth]{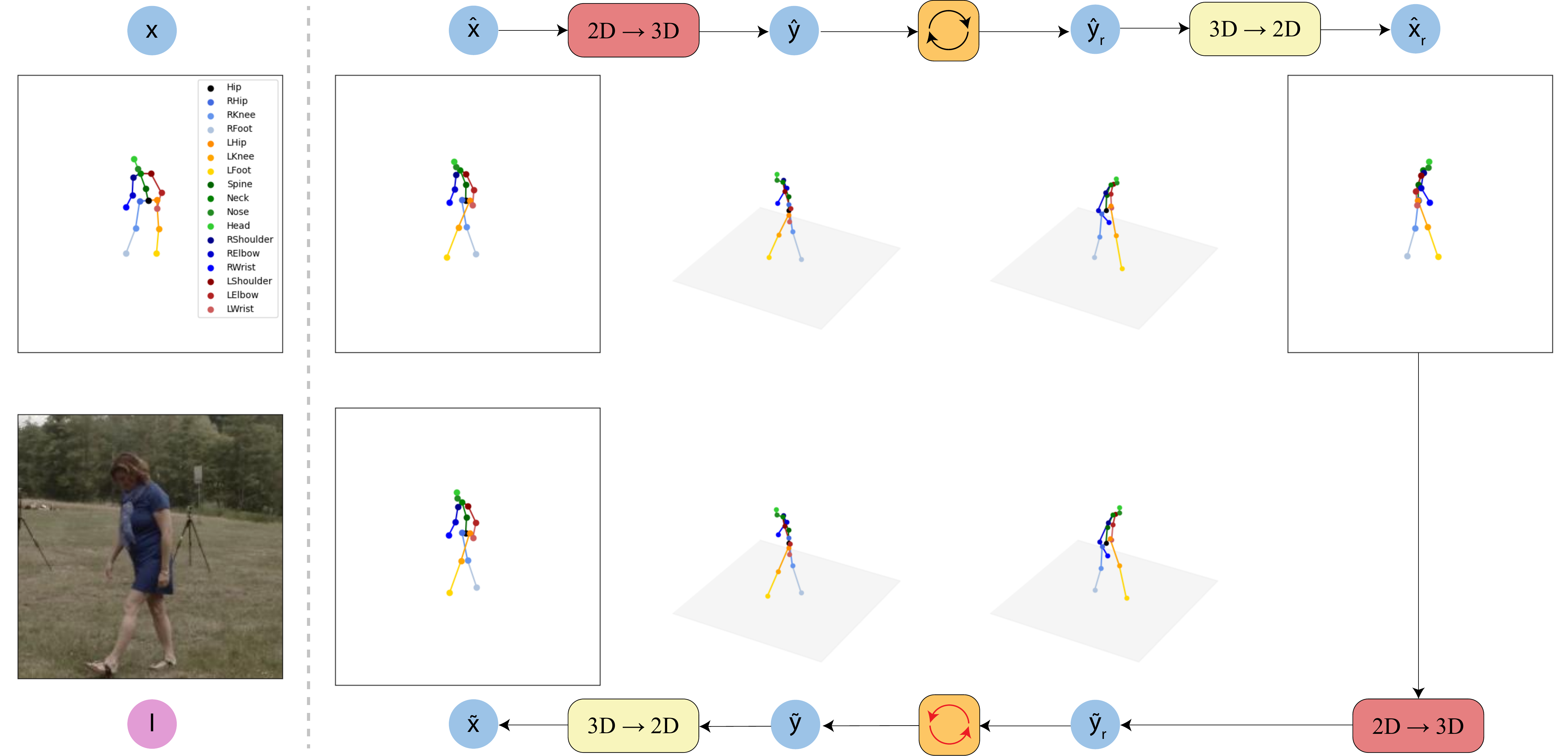}
    \caption{Additional \textbf{LiftNet} intermediate results on MPI-INF-3DHP \cite{mono-3dhp2017}. In this particular example, our 2D joint predictions $\hat{x}$ coming from the \textbf{RegNet} correct an annotation error (right and left legs are flipped) in the original annotation $x$, leading to a possibly better 3D annotation as well.}
    \label{fig:supp_res_2}
\end{figure*}

In Fig.~\ref{fig:supp_res_1} we show additional qualitative results for \textbf{LiftNet}. Our method shows promising results even when the original annotation $x$ is not correct (see Fig.~\ref{fig:supp_res_2}), since the estimated camera ($[K] [R|t]$) and the estimated 2D pose ($\hat{x}$) that \textbf{RegNet} gives as output are robust enough to provide a good pseudo-ground truth.

\clearpage


\bibliographystyle{unsrt}  
\bibliography{main}

\end{document}